\documentclass[10pt,twocolumn,letterpaper]{article}

\usepackage{cvpr}              

\usepackage{multirow}

\definecolor{cvprblue}{rgb}{0.21,0.49,0.74}
\usepackage[pagebackref,breaklinks,colorlinks,allcolors=cvprblue]{hyperref}

\def\paperID{8307} 
\def\confName{CVPR}
\def\confYear{2025}

\begin{document}
\title{Breaking the Bias: \\
Recalibrating the Attention of Industrial Anomaly Detection}

\author{
Xin Chen\textsuperscript{1}, Liujuan Cao\textsuperscript{1}\thanks{Corresponding author}, Shengchuan Zhang\textsuperscript{1}, Xiewu Zheng\textsuperscript{1}, Yan Zhang\textsuperscript{1}\\
\noindent\textsuperscript{1}Key Laboratory of Multimedia Trusted Perception and Efficient Computing, \\
Ministry of Education of China, Xiamen University\\
{\tt\small 23020231154134@stu.xmu.edu.cn, \{caoliujuan, zsc\_2016, zhengxiawu, bzhy986\}@xmu.edu.cn}
}
\maketitle
\begin{abstract}
Due to the scarcity and unpredictable nature of defect samples, industrial anomaly detection (IAD) predominantly employs unsupervised learning.
However, all unsupervised IAD methods face a common challenge: the inherent bias in normal samples, which causes models to focus on variable regions while overlooking potential defects in invariant areas. 
To effectively overcome this, it is essential to decompose and recalibrate attention, guiding the model to suppress irrelevant variations and concentrate on subtle, defect-susceptible areas.
In this paper, we propose \textbf{R}ecalibrating \textbf{A}ttention of Industrial \textbf{A}nomaly \textbf{D}etection (RAAD), a framework that systematically decomposes and recalibrates attention maps. RAAD employs a two-stage process: first, it reduces attention bias through quantization, and second, it fine-tunes defect-prone regions for improved sensitivity. Central to this framework is Hierarchical Quantization Scoring (HQS), which dynamically allocates bit-widths across layers based on their anomaly detection contributions.
HQS dynamically adjusts bit-widths based on the hierarchical nature of attention maps, compressing lower layers that produce coarse and noisy attention while preserving deeper layers with sharper, defect-focused attention. This approach optimizes both computational efficiency and the model’s sensitivity to anomalies.
We validate the effectiveness of RAAD on 32 datasets using a single 3090ti. Experiments demonstrate that RAAD, balances the complexity and expressive power of the model, enhancing its anomaly detection capability. 

\end{abstract}    

\begin{figure}[t]
  \includegraphics[width=1\linewidth]{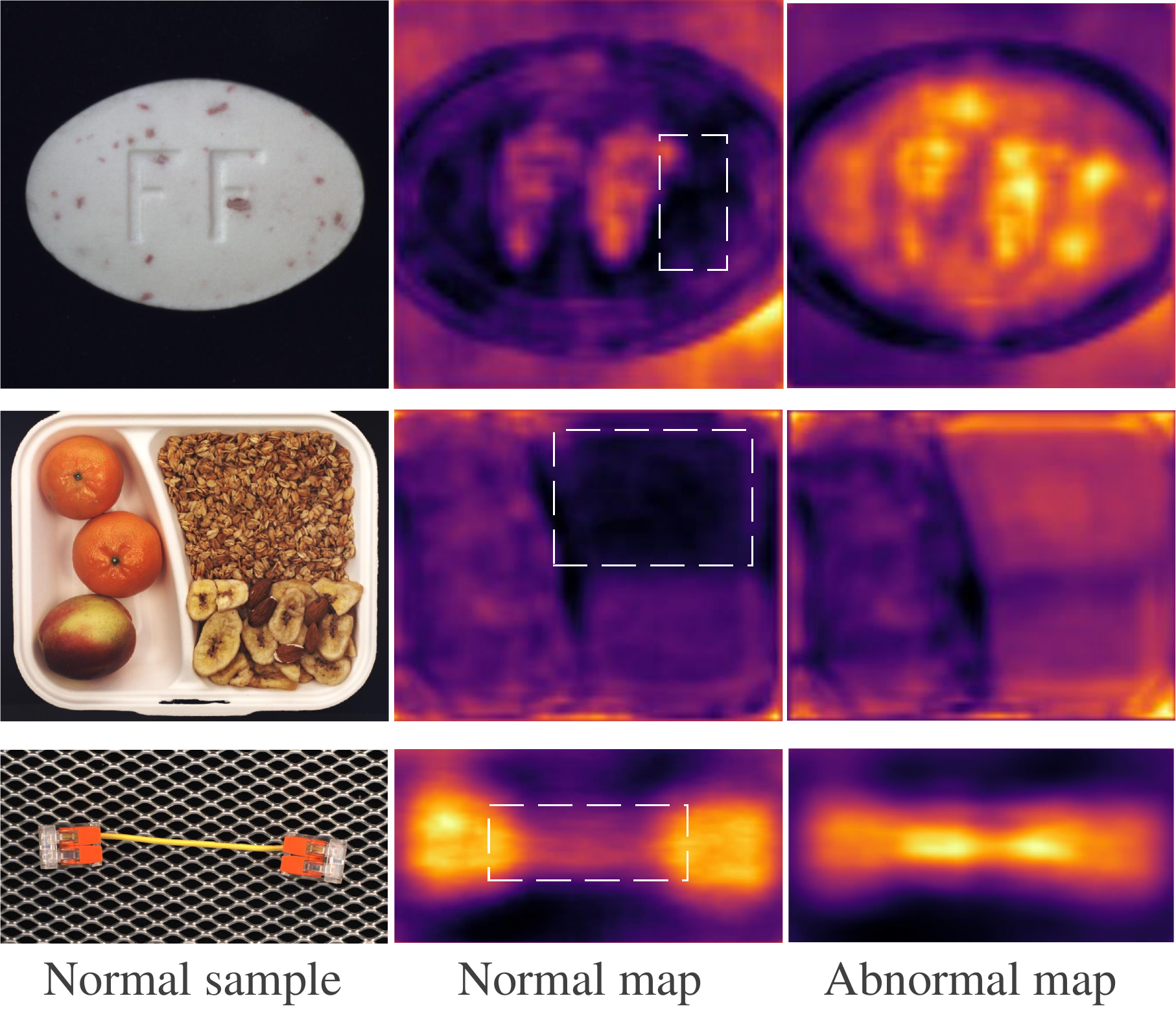}
  \caption{Visualization of heatmaps. These samples are from the MVTec-AD and MVTec LOCO datasets, which represents examples of industrial products, the average heatmap of normal samples, and the average heatmap for anomaly samples, respectively. It clearly shows the bias contained in the normal samples compared to the abnormal samples} 
  \label{fig0} 
\end{figure}

\section{Introduction}
\label{sec:intro}


Industrial anomaly detection (IAD) is crucial for maintaining the quality and safety of manufacturing processes. Because of high annotation costs and the unpredictable nature of defects, unsupervised methods have become a practical solution for real-world anomaly detection, which is only trained on normal samples.
%
%
However, traditional unsupervised methods face a fundamental challenge: during training, models tend to overfit the changing parts of normal samples while overlooking potential defects in unchanged regions. As shown in Figure~\ref{fig0}, we visualize this challenge through anomaly heatmaps on MVTec-AD~\cite{Bergmann2019MVTecAD} and MVTec LOCO~\cite{bergmann2022beyond} datasets.
Within each group, from left to right, are the normal samples of the corresponding categories, the average anomaly heatmap for normal samples, and the average anomaly heatmap for anomalous samples. Brighter areas in the heatmaps indicate regions with a higher likelihood of receiving attention.
The white boxes in the second column highlight how the model is misled by the inherent bias in normal samples.
%
%
In other words, the attention maps derived from unsupervised training tend to highlight variable regions in normal samples, thereby neglecting invariant regions where subtle anomalies may reside. One might intuitively consider abandoning the attention mechanism. Nevertheless, ignoring attention maps entirely is not a viable solution, as they play a crucial role in anomaly detection. A key question arises: how to make the model allocate attention more reasonably?

\begin{figure}
  \includegraphics[width=1\linewidth]{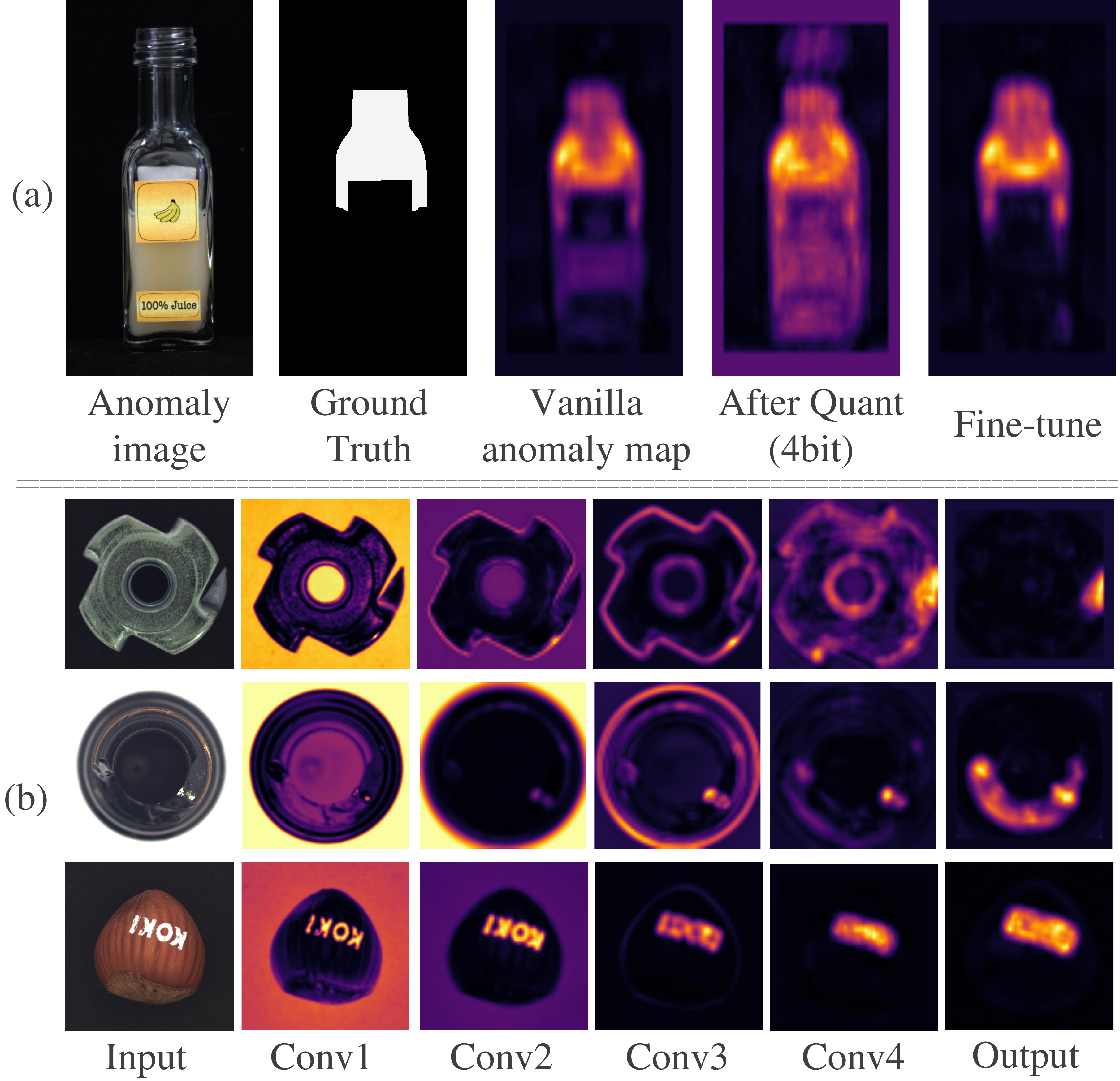}
  \caption{(a) visualization of the attention maps at different stages of the model, from left to right, are the anomaly image, ground-truth, and predicted anomaly score. (b) the layer-wise attention outputs, demonstrating the varying importance of each layer in anomaly detection. } 
  \label{fig1} 
\end{figure}

A feasible solution is to solve this problem with two steps: first, directing the model’s attention toward the primary target, and then reallocating the attention for improved anomaly detection. The former can be achieved through model quantization, while the latter is accomplished via fine-tuning.
%
During quantization, the reduction in parameter precision compels the model to prioritize learning and extracting the most critical information.
%
Meanwhile, during the fine-tuning process, the model’s attention is recalibrated, enabling the redistribution of attention to better align with task-specific requirements.
Building on this insight, we propose RAAD (\textbf{R}ecalibrating \textbf{A}ttention of Industrial \textbf{A}nomaly \textbf{D}etection), which firstly modifies attention maps with quantization and then fine-tuning them to recalibration.
%
%
Meanwhile, we observe that convolutional neural networks are commonly used as backbone networks for extracting image features in industrial anomaly detection tasks, with each layer having a different impact on the model's attention. 
To optimize the attention allocation process, we introduce \textbf{H}ierarchical \textbf{Q}uantization \textbf{S}coring (HQS), which adaptively allocates bit-width according to each layer's anomaly detection capability.
In Figure~\ref{fig1}(a), we visualize the anomaly heatmaps before and after model quantization. It can be observed that, compared to before quantization, the model's attention is more spread across the main subject while ignoring the background. Subsequently, more precise anomaly detection was achieved after fine-tuning.
Figure~\ref{fig1}(b) illustrates the outputs of each convolutional layer in the teacher-student network, highlighting the layer-wise variation in focus across the image.
This design leverages the distinct roles of network layers: shallow layers capture local details, while deeper layers extract global features, and it is most beneficial for enhancing model performance with fewer parameters.

Our main contributions are as follows:
\begin{itemize}
    \item We break the inherent bias in attention allocation within unsupervised IAD, guiding models to better detect subtle anomalies in invariant regions.
    \item We proposed RAAD, which systematically refines attention maps, using quantization to reduce bias and recalibrating the attention map via fine-tuning to improve anomaly sensitivity.
    \item We introduce HQS, a module that dynamically allocates bit-widths based on each layer's anomaly detection capability, optimizing the alignment between quantization and attention for enhanced efficiency and accuracy in IAD.
\end{itemize}

\section{Related Work}

\subsection{Unsupervised Industrial Anomaly Detection.}
Based on deep learning, visual detection has made significant achievements with the assistance of supervised learning, as cited in \cite{kwon2019survey,Ruff2021a}. However, in real-world industrial scenarios, the scarcity of defect samples, the cost of annotation, and the lack of prior knowledge about defects may render supervised methods ineffective. In recent years, unsupervised anomaly detection (IAD) algorithms have been increasingly applied to industrial detection tasks, as referenced in \cite{Sydney2019deep,xie2024imiad,tao2022deep}. ``Unsupervised'' means that the training phase only includes normal images, without any defect samples. IAD refers to the task of differentiating defective images from the majority of non-defective images at the image level.
Unsupervised IAD is mainly categorized into three types, i.e., the reconstruction-based methods, the synthesizing-based methods, and the embedding-based methods. 
Feature embedding-based methods have recently achieved state-of-the-art performance and can be specifically categorized into: teacher-student architecture \cite{Bergmann2020Uninformed,Deng2022Anomaly}, normalizing flow \cite{Rezende2015Variational,Rudolph2021same}, memory bank\cite{roth2022towards,Cohen2020subimage}, and one-class classification \cite{Sohn2020learning}.

\begin{figure}[t]
  \includegraphics[width=1\linewidth]{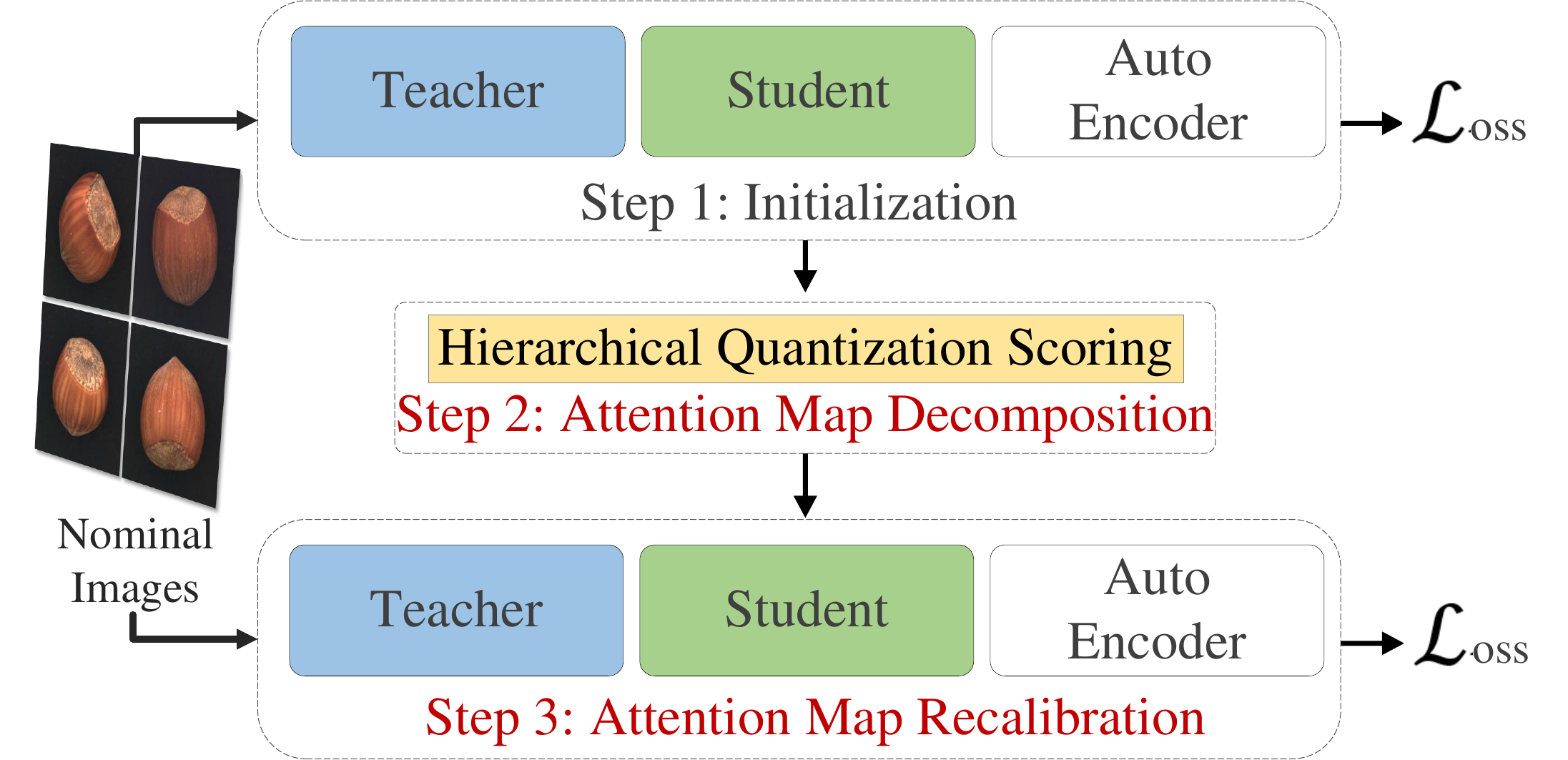}
  \caption{Pipeline of RAAD. Our architecture consists of three components: the teacher-student model and the autoencoder. During training and fine-tuning, we only use normal images. The process is divided into three steps: 1. Initial training of the model, 2. Decomposition of attention map in hierarchical quantitative scoring, detailed in Figure~\ref{layer-wise}. 3. Fine-tuning of attention recalibration.} 
  \label{pipeline} 
\end{figure}

The most typical methods are the memory bank and teacher-student architecture. 
Memory bank methods embed normal features into a compressed space. Anomalous features are distant from the normal clusters within the embedding space. 
Regarding the teacher-student architecture, the teacher is a pre-trained and frozen CNN, and the student network is trained to mimic the teacher's output on training images. Since the student has not seen any anomalous images during training, it is generally unable to predict the teacher's output on these images, thereby achieving anomaly detection.
Uninformed Students \cite{Bergmann2020Uninformed} first introduced a new framework for anomaly detection known as the teacher-student anomaly detection framework. Reverse Distillation (RD) \cite{Deng2022Anomaly} proposed a method where the student decoder learns to recover features from the compact embeddings of the teacher encoder. The GCCB \cite{zhang2024contextual} method employs a dual-student knowledge distillation framework, enhancing the ability to detect structural and logical anomalies.
However, methods based on feature embeddings rely on the size of the memory bank or the capability of the teacher network. This reliance can lead to excessive memory usage, resulting in slower inference times, or may limit the model's generalization ability.

Reconstruction-based methods \cite{Haselmann2018Anomaly, Ristea2022Self, Zavrtanik2021Reconstruction} span from autoencoders \cite{Bergmann2019Improving, Zavrtanik2021DR, chen2023easynet} and generative adversarial networks \cite{Yan2022Learning, duan2023few} to Transformers \cite{you2022unified, yao2023focus} and diffusion models \cite{lu2023removing, zhang2023unsupervised}.
Among them, autoencoder methods rely on accurately reconstructing normal images and inaccurately reconstructing anomalous ones, detecting anomalies by comparing the reconstruction with the input image. Reconstruction-based methods are more likely to capture information from the entire image \cite{Liu2020Towards}. However, these methods often produce blurry and inaccurate reconstructions, leading to an increase in false positives and generally underperforming compared to the aforementioned local methods.

Recent IAD methods explore faster inference speeds while pursuing accuracy \cite{roth2022towards, batzner2024efficientad}. To balance accuracy and inference speed, RAAD adopts the student-teacher architecture, avoiding the additional GPU memory of storing a normal feature. Simultaneously, it employs an autoencoder to analyze global information better, enhancing the model's capability at the pixel level.

\begin{figure}
  \includegraphics[width=1\linewidth]{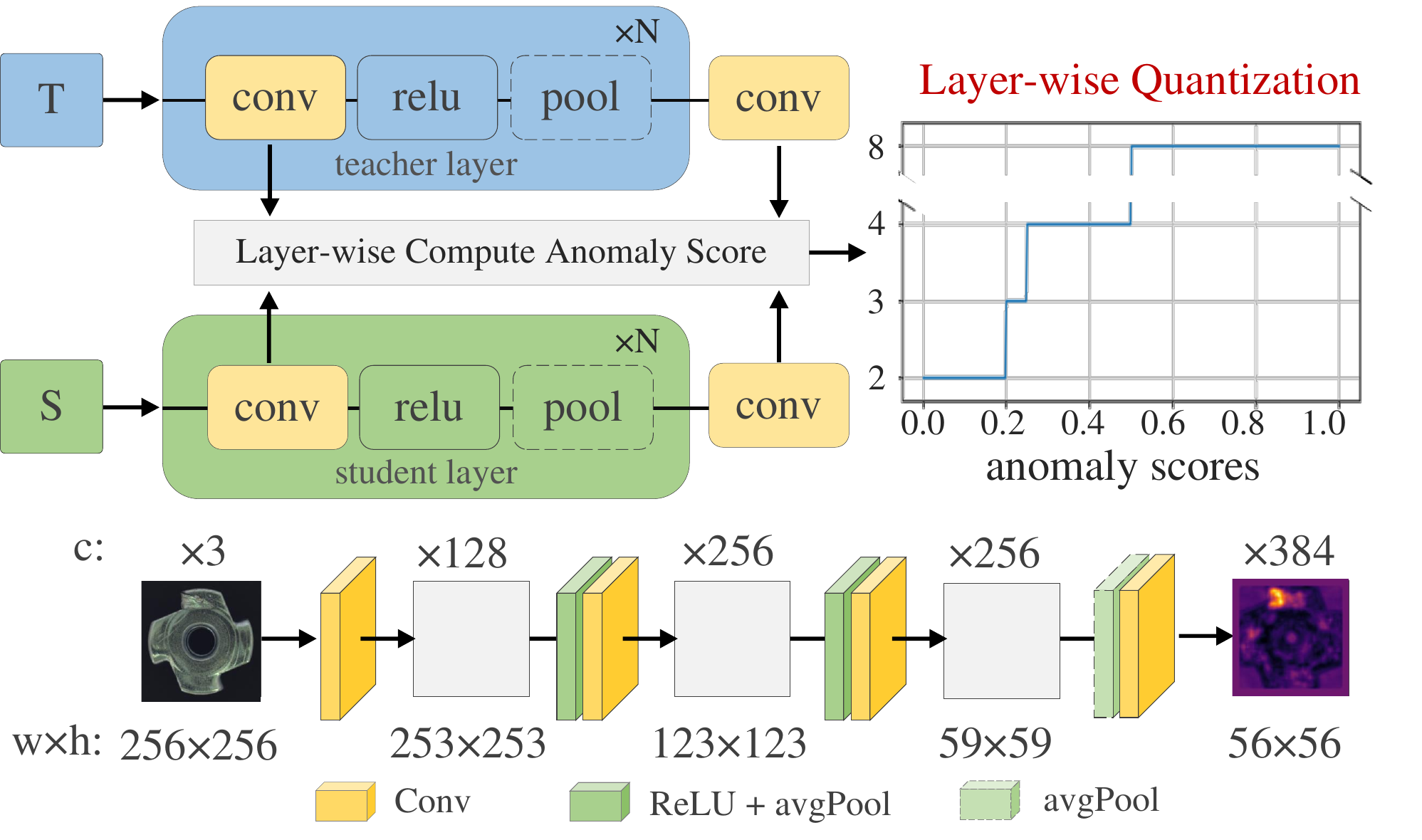}
  \caption{Hierarchical Quantization Scoring (HQS) Module. The teacher and student models are aligned layer by layer, with the anomaly scores calculated using the outputs of their respective convolutional layers. These scores are then converted into quantization bit-widths through a piecewise function. Below are the details of the teacher-student network (PDN).} 
  \label{layer-wise} 
\end{figure}

\subsection{Quantized Neural Networks.}
Quantization aims to compress models by reducing the bit precision used to represent parameters and/or activations~\cite{Cai2020ZeroQ}.
Existing neural network quantization algorithms can be divided into two categories based on their training strategy: post-training quantization (PTQ) and quantization-aware training (QAT). PTQ~\cite{Nagel2019Data} refers to quantizing the model after training, without any fine-tuning or retraining, thus allowing for quick quantization but at the cost of reduced accuracy. In contrast, QAT~\cite{gong2019differentiable,dong2020hawq} adopts an online quantization strategy. This type of method utilizes the whole training dataset during the quantization process. As a result, it has higher accuracy but limited efficiency.

Recently, several studies have explored the integration of quantization techniques into anomaly detection tasks~\cite{sharmila2023quantized}. For example,~\cite {cho2024cnn, jena2024unified} have even applied Post-Training Quantization for On-Device Anomaly Detection, striking a balance between computational efficiency and detection accuracy. However, it is important to clarify that model quantization is the method in this paper, not the goal.

%
Our method leverages the precision reduction characteristics of PTQ to achieve dimensionality reduction in weight precision, while also designing a mixed-precision quantization method specifically tailored for industrial anomaly detection.

    

\section{Method}

\subsection{Model Architecture}

Our model consists of a teacher-student model and autoencoder, as illustrated in Figure~\ref{pipeline}. The RAAD process is divided into three steps:
1. Model Initialization: The model is trained on a dataset containing only normal images. During training, only the weights of the student model and the autoencoder are updated. 2. Hierarchical Quantization Scoring: We first evaluate the anomaly detection capability of each layer in the network. Then post-training quantization of the model layer by layer. 3. Dimensionality increase: Similar to the first step, we fine-tune the student model and the autoencoder.
Before training, we employ a pre-trained WideResNet-101 (WRN-101)~\cite{Zagoruyko2016Wide} on ImageNet to initialize the teacher model. By minimizing the mean squared error(MSE) between the teacher model and the pre-trained network features. The loss function is as follows:
\begin{equation}
L_{pre} = Mean(\left\|E(I)- T(I)\right\|^2),
\end{equation}
where $I$ represents an image from the ImageNet, $E$ is a feature extractor composed of the second and third layers of the pre-trained WRN-101 network, and $T(\cdot)$ refers to the teacher model.

We utilize the Patch Description Network (PDN)\cite{batzner2024efficientad} as both the teacher and student model's feature extraction network. Unlike recent anomaly detection methods that commonly employ pre-trained CNN networks, such as DenseNet-201\cite{li2023target, huang2017densely} and WideResNet-101~\cite{esser2019learned, Zagoruyko2016Wide}, the PDN consists of only four convolutional layers. It is fully convolutional and can be applied to images of variable sizes. As the network depth and parameters are reduced, the running time and memory requirements of the model are correspondingly reduced, achieving an inference speed of 113 FPS for a single image on an NVIDIA RTX3090 GPU.

While PDN has the advantage of focusing on patch features, this can be detrimental for anomaly localization in logical anomaly detection tasks that require global information. As suggested by the authors of the MVTec LOCO-AD dataset~\cite{Bergmann2021MVTecAD}, we use an autoencoder for learning logical constraints of the training images and detecting violations of these constraints, comprising stridden convolutions in the encoder and bilinear upsampling in the decoder. Compared to the patch-based student model, the autoencoder encodes and decodes the complete image through a bottleneck of 64 latent dimensions.

\begin{figure}[t!]
  \includegraphics[width=1\linewidth]{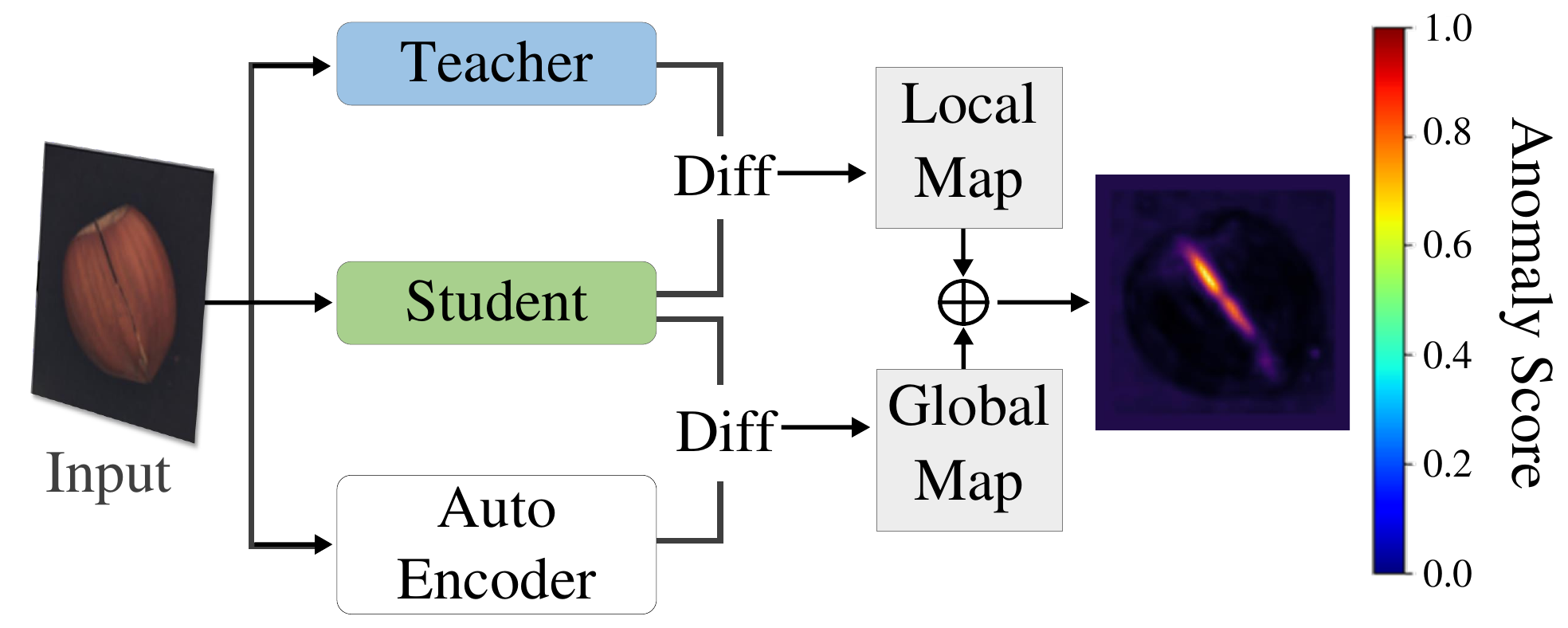}
  \caption{The inference process of the models, the input is from the MVTec AD test dataset. ``Diff'' refers to computing the element-wise squared difference between two collections of output feature maps and computing its average across feature maps. To obtain pixel anomaly scores, the anomaly maps are resized to match the input image using bilinear interpolation.} 
  \label{infere} 
\end{figure}

\subsection{Attention Map Decomposition}

Our quantization method refers to BRECQ~\cite{Li2021BRECQ}, and compared to other post-training quantization methods, it adopts a block-wise reconstruction strategy instead of the traditional layer-wise or network-wise reconstruction. Block-wise reconstruction takes into account the dependencies within the block while ignoring the dependencies between blocks, which has significant advantages for patch-based networks.
Using the diagonal Fisher Information Matrix (FIM), we measure the cross-layer dependencies within each block and convert the second-order error of any block into the output of that block: $\mathbb{E}\left[\Delta \mathbf{z}^{(\ell), \mathrm{T}} \mathbf{H}^{\left(\mathbf{z}^{(\ell)}\right)} \Delta \mathbf{z}^{(\ell)}\right]$.
The output of the neural network is $z^{\left(n\right)}=f\left(\theta\right)$, $\mathbf{z}^{\left(n\right)} \in \mathcal{R}^{m}$. $\mathbf{H}^{ \left(\cdot \right) }$ represents the diagonal Hessian of the intermediate block. If layers $k$ to $\ell$ $\left(1 \leq k \leq \ell \leq n\right)$ form a block, the weight vector is defined as $\tilde{\theta} = \operatorname{vec}\left[\mathbf{w}^{\left(k\right), \mathbf{T}}, \ldots, \mathbf{w}^{\left(\ell\right), \mathbf{T}}\right]^{\top}$, which is the concatenated vector of the weights from layer $k$ to layer $\ell$.

Formally, given a probability model $p(x \mid \theta)$, the FIM is equal to the negative expected Hessian matrix of the log-likelihood function, defined as:
\begin{equation}
\begin{aligned}
\overline{\mathbf{F}}^{(\theta)}&=\mathbb{E}\left[\nabla_{\theta} \log p_{\theta}(y \mid x) \nabla_{\theta} \log p_{\theta}(y \mid x)^{\top}\right]\\
&=-\mathbb{E}\left[\nabla_{\theta}^{2} \log p_{\theta}(y \mid x)\right]=-\overline{\mathbf{H}}_{\log p(x \mid \theta)}^{(\theta)}.
\end{aligned}
\end{equation}

The diagonal of the pre-activation FIM is equal to the squared gradient of each element, with the optimization objective being:
\begin{equation}
\min _{\hat{\mathbf{w}}} \mathbb{E}\left[\Delta \mathbf{z}^{(\ell), \mathrm{T}} \operatorname{diag}\left(\left(\frac{\partial L}{\partial \mathbf{z}_{1}^{(\ell)}}\right)^{2}, \ldots,\left(\frac{\partial L}{\partial \mathbf{z}_{a}^{(\ell)}}\right)^{2}\right) 
\Delta \mathbf{z}^{(\ell)}\right].
\end{equation}

\subsection{Hierarchical Quantization Scoring}
To further enhance the redistribution of attention in anomaly detection, we propose a Hierarchical Quantization Scoring (HQS) mechanism, which utilizes mixed-precision techniques to adaptively adjust the bit-width for each layer. As illustrated in Figure~\ref{layer-wise}, both the teacher and student models are processed through the HQS module, where corresponding layers of the same depth are aligned, and their outputs after convolution are used to compute an anomaly score $s$ ($s \in (0,1)$). This score measures the alignment of attention between the two models, guiding the reallocation of attention to defect-prone regions.

The bit-width $b$ for the $\ell$-th layer ($1 \leq \ell \leq N$) is determined as a function of its anomaly score. Layers with higher scores, which contribute more to accurate anomaly localization, are assigned greater precision, allowing deeper, semantically rich layers to focus on subtle anomalies. In contrast, shallower layers, which often capture redundant or noisy information, are more heavily compressed. This dynamic bit-width allocation leads to a more effective redistribution of attention, ensuring that the model emphasizes critical regions in the attention maps.
%
%
The bit width $b$ for the $\ell$-th layer ($1 \leq \ell \leq N$) is defined as:
\begin{equation}
\begin{aligned}
b^{(\ell)}&=\phi(anomaly\_score)\\
&=\phi\left((c^{(\ell)}w^{(\ell)}h^{(\ell)})^{-1}\sum_{c}\left\|\mathrm{T}^{(\ell)}_{c}(i)-\mathrm{S}^{(\ell)}_{c}(i)\right\|_{F}^{2}\right),
\end{aligned}
\end{equation}
where $i$ represents the output of the previous layer, when $\ell = 1$, $i = conv^{(1)}(I)$, with $I$ being the image input to the model. $T(i), S(i) \in \mathbb{R}^{c \times w \times h}$, where $c, w, h$ are the number of channels, width, and height of the output features of the $\ell$-th layer, respectively. $\phi(\cdot)$ is a piecewise function that determines the bit width based on the hierarchical quantization score. We chose 2, 3, 4, and 8 bits for mixed precision because they are most common in practical deployment.

\subsection{Attention Map Recalibration}

During the training process, the teacher model, student model, and autoencoder are paired with each other to generate three losses: $L_{t-s}, L_{ae-s}, L_{t-ae}$. Formally, we apply the teacher $T$, student $S$, and autoencoder $A$ to the training image $I$, with $T(I), S(I), A(I) \in \mathbb{R}^{C \times W \times H}$, and the loss expression for $L_{t-s}$ is:
\begin{equation}
L_{\mathrm{t-s}} = (C W H)^{-1} \sum_{c}\left\|T(I)_{c}-S(I)_{c}\right\|_{F}^{2},
\end{equation}
The expressions for $L_{ae-s}$ and $L_{t-ae}$ are similar to $L_{t-s}$, differing in that $T\left(I\right)_{c}-S\left(I\right)_{c}$ is replaced with $A\left(I\right)_{c}-S\left(I\right)_{c}$ and $T\left(I\right)_{c}-A\left(I\right)_{c}$, respectively.
Note that to confine $L_{t-s}$ to the most relevant parts of the image, the value of 10\% is used for backpropagation in each of the three dimensions of the mean squared error $D$, where $ D_{c,w,} = \left(T \left(I\right)_{c,w,h} - S\left(I\right)_{c,w,h}\right)^{2}$.

The total loss is the weighted summation of the three: 
\begin{equation}
Loss=\lambda_{t-s}L_{t-s}+\lambda_{ae-s}L_{ae-s}+\lambda_{t-ae}L_{t-ae}.
\end{equation}
As illustrated in Figure~\ref{infere}, the inference process after training involves the teacher-student outputs a local anomaly map, while the autoencoder-student outputs a global anomaly map. These two anomaly maps are averaged to calculate a composite anomaly map, with its maximum value used as the image-level anomaly score, where the $2D$ anomaly score map $M \in R^{W \times H}$ is given by $M_{w,h} = C^{-1}\sum_c D_{c,w,h}$, which is the cross-channel average of $D$, assigning an anomaly score to each feature vector.

\begin{table*}[t!]
    \begin{center}
    \setlength\tabcolsep{5pt}
    \begin{tabular}{l|ccccc}
    \toprule[1.0pt]
        \textbf{Dataset}& Baseline$^{\dagger}$& LSQ& OMPQ& RAAD\\ 
        \midrule[0.5pt]
        \textbf{W/A}& 32/32 & 8/8 & 8/8 & $\leq$8/$\leq$8 \\
        \midrule[0.5pt]\midrule[0.5pt]
        \textbf{MVTec AD}& 96.98 $\backslash$ 97.44 $\backslash$ 91.38& 
        97.21 $\backslash$ 96.58 $\backslash$ 86.19& 
        98.77 $\backslash$ 97.97 $\backslash$ 92.17& 
        \textbf{98.90 $\backslash$ 97.83 $\backslash$ 92.92}& \\
        \midrule[0.5pt]
        bottle& \textbf{100.0 $\backslash$ 100.0 $\backslash$ 94.58}& 
        99.92 $\backslash$ 98.32 $\backslash$ 88.82 & 
        \textbf{100.0 $\backslash$ 100.0} $\backslash$ 93.92& 
        \textbf{100.0 $\backslash$ 100.0} $\backslash$ 93.97& \\
        cable&  95.16$\backslash$ \textbf{100.0} $\backslash$ \textbf{94.58}&  
        95.25 $\backslash$ 86.34 $\backslash$ 87.39& 
        96.49 $\backslash$ 90.72 $\backslash$ 86.80&
        \textbf{97.71} $\backslash$ 94.51 $\backslash$ 88.75& \\
        capsule& 94.41 $\backslash$ 94.32 $\backslash$ 96.00& 
        85.32 $\backslash$ 87.23 $\backslash$ 79.17& 
        96.33 $\backslash$ \textbf{96.40 $\backslash$ 96.78}& 
        \textbf{97.40} $\backslash$ 94.74 $\backslash$ 96.70& \\
        carpet& 97.43  $\backslash$ 98.91 $\backslash$ 91.11& 
        98.17 $\backslash$ 98.51 $\backslash$ 90.38& 
        98.71 $\backslash$ 98.84 $\backslash$ 91.09& 
        \textbf{98.79 $\backslash$ 98.85 $\backslash$ 91.99}& \\
        grid& 99.08 $\backslash$ 100.0 $\backslash$ 88.84& 
        \textbf{100.0 $\backslash$ 100.0} $\backslash$ 88.85& 
        \textbf{100.0 $\backslash$ 100.0} $\backslash$ 88.75& 
        99.83 $\backslash$ 100.0 $\backslash$ \textbf{91.00}& \\
        hazelnut& 99.50 $\backslash$ \textbf{100.0} $\backslash$ 91.40& 
        99.14 $\backslash$ 99.18 $\backslash$ 83.88& 
        99.50 $\backslash$ \textbf{100.0 $\backslash$ 92.44}& 
        \textbf{99.78 $\backslash$ 100.0 $\backslash$ 92.44}& \\
        leather& 86.68 $\backslash$ 93.23 $\backslash$ 97.09& 
        98.30 $\backslash$ 97.20 $\backslash$ 97.87& 
        \textbf{99.79 $\backslash$ 98.92 $\backslash$ 98.22}& 
        98.30 $\backslash$ 93.88 $\backslash$ 97.87& \\
        metal$\_$nut& 98.43 $\backslash$ 98.76 $\backslash$ 91.86& 
        97.99 $\backslash$ 98.24 $\backslash$ 89.09& 
        98.77 $\backslash$ \textbf{98.91} $\backslash$ 93.28& 
        \textbf{98.82 $\backslash$ 98.91 $\backslash$ 93.66}& \\
        pill& 96.78 $\backslash$ \textbf{98.34} $\backslash$ 95.87& 
        92.03 $\backslash$ 95.33 $\backslash$ 84.14& 
        97.84 $\backslash$ 97.18 $\backslash$ 97.41& 
        \textbf{98.00} $\backslash$ 97.18 $\backslash$ \textbf{97.44}& \\
        screw& 93.72 $\backslash$ 93.44 $\backslash$ 89.87& 
        96.67 $\backslash$ 96.55 $\backslash$ 90.96& 
        98.48 $\backslash$ 95.08 $\backslash$ \textbf{94.97}& 
        \textbf{98.56 $\backslash$ 96.64} $\backslash$ 94.45& \\
        tile& \textbf{100.0 $\backslash$ 100.0} $\backslash$ 88.42& 
        \textbf{100.0 $\backslash$ 100.0} $\backslash$ 88.42& 
        \textbf{100.0 $\backslash$ 100.0} $\backslash$ 88.42& 
        \textbf{100.0 $\backslash$ 100.0} $\backslash$ \textbf{89.00}& \\
        toothbrush& \textbf{100.0 $\backslash$ 100.0 $\backslash$ 94.47}& 
        \textbf{100.0 $\backslash$ 100.0} $\backslash$ 58.72& 
        \textbf{100.0 $\backslash$ 100.0 $\backslash$ 94.47}& 
        \textbf{100.0 $\backslash$ 100.0 $\backslash$ 94.47}&\\
        transistor& 99.54 $\backslash$ \textbf{100.0} $\backslash$ 85.43& 
        99.29 $\backslash$ \textbf{100.0} $\backslash$ 85.43& 
        \textbf{100.0 $\backslash$ 100.0 $\backslash$ 87.31}& 
        \textbf{100.0 $\backslash$ 100.0 $\backslash$ 87.31}&\\
        wood& 98.77 $\backslash$ 97.63 $\backslash$ 87.41& 
        \textbf{99.03} $\backslash$ 97.69 $\backslash$ \textbf{87.57}& 
        98.68 $\backslash$ \textbf{98.36} $\backslash$ 86.65& 
        98.50 $\backslash$ \textbf{98.36} $\backslash$ 87.00& \\
        zipper& 95.24 $\backslash$ 94.43 $\backslash$ 91.67& 
        97.05 $\backslash$ 94.18 $\backslash$ 92.15& 
        96.95 $\backslash$ \textbf{95.16} $\backslash$ 92.05& 
        \textbf{97.84} $\backslash$ 94.44 $\backslash$ \textbf{97.84}& \\ 
        \midrule[0.5pt]\midrule[0.5pt]
        \textbf{LOCO}& 84.09$\backslash$ 78.51 $\backslash$ 83.32& 
        86.26 $\backslash$ 82.34 $\backslash$ 81.16& 
        89.60 $\backslash$ \textbf{88.59} $\backslash$ 86.19& 
        \textbf{89.75} $\backslash$ 87.85 $\backslash$ \textbf{86.76}& \\
        \midrule[0.5pt]
        breakfast$\_$box& 77.13$\backslash$ 66.98 $\backslash$ 65.21& 
        80.54 $\backslash$ 66.93 $\backslash$ 65.31& 
        80.91 $\backslash$ \textbf{91.79} $\backslash$ 72.00& 
        \textbf{80.94} $\backslash$ 85.81 $\backslash$ \textbf{72.31}& \\ 
        juice$\_$bottle& 96.41$\backslash$ 96.79 $\backslash$ 97.28& 
        \textbf{99.62 $\backslash$ 98.72 $\backslash$ 98.29}& 
        97.86 $\backslash$ 97.41 $\backslash$ 97.96& 
        98.21 $\backslash$ 97.01 $\backslash$ 98.07& \\
        pushpins& 78.35$\backslash$ 68.12 $\backslash$ 88.23& 
        78.97 $\backslash$ 85.03 $\backslash$ 83.47& 
        95.46 $\backslash$ 90.59 $\backslash$ 90.86& 
        \textbf{95.85 $\backslash$ 91.18 $\backslash$ 91.09}& \\
        screw$\_$bag& 71.57$\backslash$ 67.80 $\backslash$ 76.13& 
        74.65 $\backslash$ 66.25 $\backslash$ 64.25& 
        \textbf{76.02 $\backslash$ 67.86} $\backslash$ 75.41& 
        75.95 $\backslash$ 66.36 $\backslash$ \textbf{77.64}& \\
        splicing$\_$connectors& 97.03$\backslash$ 78.51 $\backslash$ 94.53& 
        97.56 $\backslash$ 94.76 $\backslash$ 94.50& 
        97.75 $\backslash$ 95.29 $\backslash$ \textbf{94.74}& 
        \textbf{97.83 $\backslash$ 98.89} $\backslash$ 94.72& \\
        \midrule[0.5pt]\midrule[0.5pt]
        \textbf{VisA}& 94.73$\backslash$ 88.79& 94.73$\backslash$90.14 & \textbf{96.73$\backslash$92.07} & 96.72$\backslash$92.05 &\\
        \midrule[0.5pt]
        candle& 91.72$\backslash$ 81.90 & 85.85$\backslash$75.68 & 
        \textbf{93.04} $\backslash$86.60 & 92.92$\backslash$\textbf{87.50} &\\
        capsules& 85.15$\backslash$ 76.03& 83.16$\backslash$73.02 & 
        \textbf{86.15} $\backslash$ 77.97& 85.13$\backslash$\textbf{79.46} &\\
        cashew& 86.15$\backslash$ 76.03& 98.32$\backslash$94.23 & 
        \textbf{98.92} $\backslash$ 94.23& 98.14$\backslash$\textbf{95.05} &\\
        chewinggum& 98.68$\backslash$ 96.08& 98.57$\backslash$97.94 & 
        \textbf{99.74 $\backslash$ 98.99}& 99.64$\backslash$ \textbf{98.99}&\\
        fryum& 97.72$\backslash$ 95.88& 97.08$\backslash$\textbf{97.83} & 
        98.40 $\backslash$ 93.33& \textbf{98.61}$\backslash$96.91 &\\
        macaroni1& 96.54$\backslash$ 87.85& 94.33$\backslash$85.19 & 
        98.20 $\backslash$ \textbf{93.07}& \textbf{98.48}$\backslash$86.84 &\\
        macaroni2& 89.26$\backslash$ 78.26& 85.89$\backslash$78.64 & 
        90.78 $\backslash$ \textbf{83.81}& \textbf{92.05}$\backslash$82.57 &\\
        pcb1& 99.17$\backslash$ 96.08& \textbf{99.58}$\backslash$97.06 & 
        99.51 $\backslash$ 95.19& 99.53$\backslash$\textbf{97.98} &\\
        pcb2& 98.94$\backslash$ 95.92& \textbf{99.36}$\backslash$96.04 & 
        99.17 $\backslash$ 92.52& 99.28$\backslash$\textbf{96.08} &\\
        pcb3& 95.71$\backslash$ 89.11& \textbf{97.79$\backslash$94.79} & 
        97.43 $\backslash$ 93.07& 97.56$\backslash$92.93 &\\
        pcb4& 99.35$\backslash$ 94.34& 99.39$\backslash$97.06 & 
        \textbf{99.64 $\backslash$ 97.09}& 99.60$\backslash$96.15 &\\
        pipe$\_$fryum& 99.54$\backslash$ 98.00& 97.50$\backslash$94.23 & \textbf{99.84$\backslash$ 99.00}& 99.72$\backslash$\textbf{99.00} &\\
        \bottomrule[1.0pt]
    \end{tabular}
    \caption{
    Mean Anomaly Detection AU-ROC/AP/Segmentation AU-PRO on MVTec AD. Mean Anomaly Detection AU-ROC(logical and structural anomalies)/AP/Segmentation AU-PRO(logical and structural anomalies) on MVTec LOCO-AD.
    Mean Anomaly Detection AU-ROC/AP on VisA. $\dagger$ denotes the unofficial implementation of EfficientAD. $\ast$ indicates mixed precision.
    } 
    \label{tab1_compare}
    \end{center}
\end{table*}


\section{Experiment}
In this section, we demonstrate the effectiveness of RAAD, by comparing the impact of different quantization methods on the performance of the model and comparing our proposed method with other advanced IAD methods. Moreover, we provide additional ablation studies.

\begin{table*}[t]
\renewcommand{\arraystretch}{1.2}
\setlength{\tabcolsep}{9pt}
    \begin{tabular}{c|c|cccc|c}
    \toprule[1.0pt]
     & Method & \begin{tabular}[c]{@{}c@{}}PatchCore\\ (CVPR2022)\end{tabular} & \begin{tabular}[c]{@{}c@{}}GCAD\\ (IJCV2022)\end{tabular} & \begin{tabular}[c]{@{}c@{}}SimpleNet\\ (CVPR2023)\end{tabular} & \begin{tabular}[c]{@{}c@{}}EfficientAD-S\\ (WACV2024)\end{tabular} & RAAD \\ \hline
    \multirow{2}{*}{\textbf{MVTecAD}} & Det. AU-ROC & 99.2 & 89.1 & \textbf{99.6} & 96.98 & 98.9 \\
     & Seg. AU-PRO & 93.5 & -- & \textbf{98.1} & 91.38 & 92.92 \\ \hline
    \multirow{3}{*}{\textbf{LOCO}} & Mean & 80.3 & 83.3 &77.6 & 84.09 & \textbf{89.75} \\
     & Logic. & 75.8 & 83.9 & 71.5 & 79.88 & \textbf{87.46} \\
     & Struct. & 84.8 & 82.7 & 83.7 & 90.12 & \textbf{91.93} \\ \hline
    \textbf{VisA} & Det. AU-ROC & -- & 89.1 & 87.9 & 95.39 & \textbf{96.72} \\
    \bottomrule[1.0pt]
    \end{tabular}
    \caption{
    Comparison with Some state-of-the-art methods. Mean Anomaly Detection AU-ROC and Segmentation AU-PRO on MVTec AD. Mean Anomaly Detection AU-ROC, logical anomalies, and structural anomalies on MVTec LOCO-AD. Mean Anomaly Detection AU-ROC on VisA.
    } 
    \label{tab2_compare_sota}
\end{table*}

\begin{table}[t]
\centering
\renewcommand{\arraystretch}{1.0}
\setlength\tabcolsep{9pt}
\begin{tabular}{cc|cc}
 \toprule[1.0pt]
  Quant & HQS & Det. AU-ROC & Seg. AU-PRO  \\ 
  \midrule[0.5pt]
   &  & 96.98 & 91.38\\
  \checkmark &  & 98.77$( \textcolor{blue}{1.79\uparrow})$ & 92.17$(\textcolor{blue}{0.79\uparrow})$ \\
  \checkmark  & \checkmark & 98.9$( \textcolor{blue}{1.92\uparrow})$ & 92.92$( \textcolor{blue}{1.54\uparrow})$ \\
    
 \bottomrule[1.0pt] 
 \end{tabular}
 \caption{
  Ablation studies on our RAAD. ``Quant'': The model utilizes post-training quantization, where the weights and activations are quantized to 8-bit precision followed by fine-tuning. ``HQS'': Using Layer-wise mixed precision Quantization. Improvements over the baseline are highlighted in \textcolor{blue}{blue}.} 
 \label{tab3_ablation_components}
\end{table}

\subsection{Datasets and Evaluation Metric} 
\textbf{MVTec AD}\cite{Bergmann2019MVTecAD} dataset is a widely recognized anomaly detection benchmark, that encompasses a diverse dataset of 5,354 high-resolution images from various domains. 
The data is divided into training and testing sets, with the training set containing 3,629 anomaly-free images, ensuring a focus on normal samples. On the other hand, the test set consists of 1,725 images, providing a mix of both normal and abnormal samples for comprehensive evaluation. To aid in the anomaly localization evaluation, pixel-level annotations are provided.
\textbf{MVTec LOCO} \cite{bergmann2022beyond} dataset includes both structural and logical anomalies. It contains 3644 images from five different categories inspired by real-world industrial inspection scenarios. 
Structural anomalies appear as scratches, dents, or contaminations in the manufactured products. Logical anomalies violate underlying constraints, e.g., a permissible object being present in an invalid location or a required object not being present at all. 
%
\textbf{VisA} dataset \cite{zou2022spot} proposes multi-instance IAD, comprising 10,821 high-resolution images, including 9,621 normal images and 1,200 anomaly images. This dataset is organized into 12 unique object classes. These 12 object classes can be further categorized into three distinct object types: Complex Structures, Multiple Instances, and Single Instances.

\textbf{Evaluation Metric.} 
We use several standard evaluation metrics, for anomaly detection, including the Area Under the Receiver Operating Characteristic Curve (Det. AUROC), and Average Precision (AP). For localization, we use the Per-Region-Overlap (PRO) metric, setting the false positive rate(FPR) of 30\%, as recommended by~\cite{Bergmann2021MVTecAD}. For MVTec LOCO, we use the AU-sPRO metric~\cite{bergmann2022beyond}, a generalization of AU-PRO  for evaluating the localization of logical anomalies. 

\subsection{Implementation Details}
We pre-train the teacher model using the pre-trained WideResnet101~\cite{Zagoruyko2016Wide} on the ImageNet dataset. Both the teacher and student models use the small version of the Patch Description Network (PDN), with the student model's output feature dimension twice that of the teacher model, and the autoencoder encodes and decodes the complete image through a bottleneck of 64 latent dimensions. 
Our hyperparameter settings are as follows. $\lambda_{t-s}$, $\lambda_{t-ae}$, $\lambda_{ae-s}$ are all set to 1. 
During both training and fine-tuning, the teacher model is frozen.
The Adam optimizer is used with a learning rate of 0.00001 for the student model and the autoencoder.
Experiments are conducted by default on an NVIDIA Geforce GTX 3090Ti with 24 GB of RAM. We train our model for 70k iterations, with a maximum of 60k iterations for fine-tuning training. However, our experiments show that the model often achieves the best performance before reaching the full 60,000 iterations.
We compared our quantization method with HQS~\cite{esser2019learned} and OMPQ~\cite{ma2023ompq}.
We fix the weight and activation of the first and last layer at 8 bits following previous works, where the search space is 2-8 bits.
Note that during comparative experiments, we disabled OMPQ's mixed precision method.


\subsection{Main Results}
We choose the architecture of EfficientAD~\cite{batzner2024efficientad} as our benchmark for evaluating our method. EfficientAD is an unsupervised IAD method utilizing a lightweight feature extractor, achieving both low error rates and high computational efficiency. We believe that the trade-off between accuracy and speed is the direction for future IAD development. It is worth noting that EfficientAD has not publicly released its code, so we reproduced their method and refer to it as baseline$^{\dagger}$ in Table~\ref{tab1_compare}. 
\begin{figure}
 \includegraphics[width=1\linewidth]{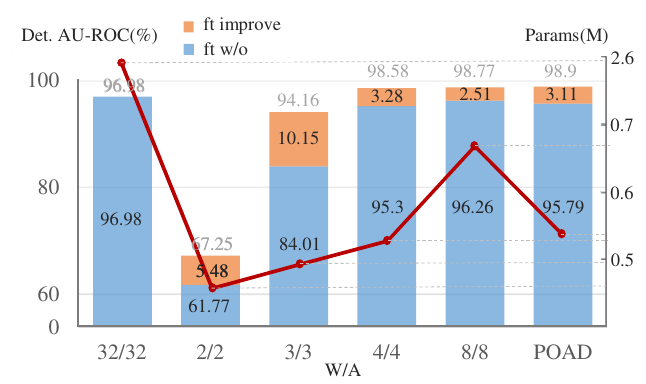}
 \caption{Quantizing and fine-tuning models using varying bit-width. Constructing stacked bar charts with Det. AU-ROC corresponds to the left vertical axis. Plotting line graphs with model parameters corresponding to the right vertical axis.} 
 \label{bar} 
\end{figure}
\begin{figure*}[t!]
 \includegraphics[width=1\linewidth]{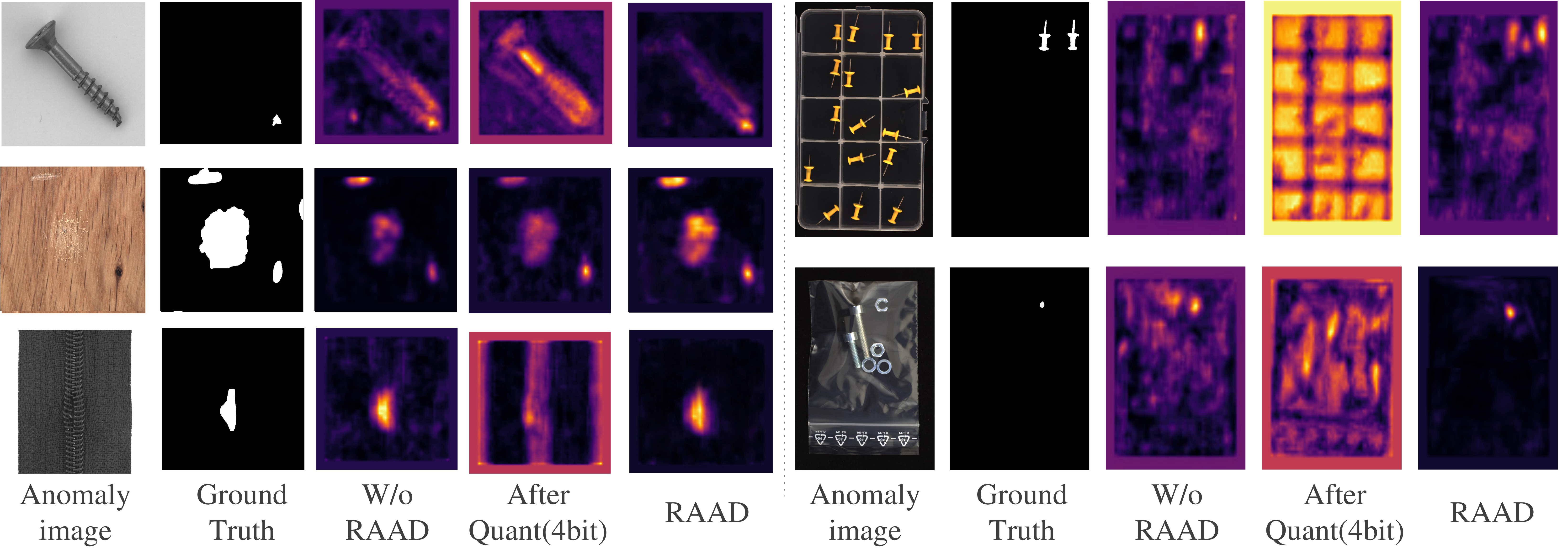}
 \caption{Qualitative results of RAAD on the MVTec-AD dataset~\cite{Bergmann2019MVTecAD} and MVTec-AD LOCO~\cite{bergmann2022beyond}. Within each group, from left to right, are the abnormal image, true value, baseline method prediction, quantized model prediction, RAAD prediction.} 
 \label{fig_vis} 
\end{figure*}
To demonstrate the improvement of RAAD on the model's anomaly detection performance, we conducted extensive experiments on 32 datasets across three IAD datasets. We applied various post-training quantization methods to the original framework for comparison, with all quantization methods performing maximum quantization, i.e., both weights and activations are quantized to 8 bits, ensuring the bit-width is greater than or equal to that used by RAAD quantization. As shown in Table~\ref{tab1_compare}, RAAD achieved average detection AU-ROC scores of 98.8, 89.75, and 96.13 on MVTec AD, MVTec LOCO, and VisA, respectively.
%
RAAD consistently outperforms baselines, emphasizing its effectiveness in both image-level anomaly detection and pixel-level anomaly localization. This demonstrates the framework’s adaptability to diverse anomaly characteristics.

In addition, we also compared RAAD with several competitive methods across multiple datasets using various evaluation metrics, as shown in Table~\ref{tab2_compare_sota}. We compare RAAD with PatchCore~\cite{roth2022towards}, GCAD~\cite{bergmann2022beyond}, and SimpleNet~\cite{liu2023simplenet}. Besides EfficientAD, the results of the other methods are from ``paper with code''. RAAD's average Detection AU-ROC score across the three datasets is 95.12, which is 5.37, 7.95, 6.75, and 2.97 higher than the other methods, demonstrating the best overall anomaly detection performance, proving RAAD achieves powerful image-level detection and pixel-level anomaly localization.


\subsection{Empirical Studies}
%

\textbf{Effectiveness of different components of RAAD.}
We investigate the effectiveness of each component of RAAD in Table~\ref{tab3_ablation_components}. We set the baseline as the performance of EfficientAD-S on MVTec AD, comparing the Mean Detection AU-ROC and Mean Segmentation AU-PRO.
Utilizing post-training quantization leads to improvements of 1.79\% and 0.79\%, respectively.
Incorporating Hierarchical Quantization Scoring (HQS) results in improvements of 0.13\% and 0.75\% respectively, indicating that the introduction of HQS can further enhance the model performance.
Quantization alone improves performance by mitigating attention bias, but HQS significantly optimizes attention distribution, emphasizing its critical role in refining anomaly sensitivity.

\textbf{Quantization and Hierarchical Quantization Scoring (HQS).}
Figure~\ref{bar} illustrates the effect of varying quantization bit-widths on model performance, including fine-tuning results for different configurations. Constructing stacked bar charts with Det. AU-ROC corresponds to the left vertical axis. Plotting line graphs with model parameters corresponding to the right vertical axis.
It can be seen that wider bit-widths better preserve the original model performance, while lower bit-widths increase precision loss, posing challenges for fine-tuning to recover dimensionality. However, it is also observed that fine-tuning is crucial for improving the performance of the quantized model, especially when using lower bit-width quantization, which can lead to more significant improvement. 
%
%
Notably, after applying the HQS method, where the quantization width does not exceed 8 bits, the performance before fine-tuning is slightly lower than that of 8-bit quantization. However, after fine-tuning, it surpasses 8-bit quantization, demonstrating that a more suitable bit-width can better preserve key correspondences.
At the same time, the line chart in Figure~\ref{bar} visualizes the model parameters. Compared to other results with fixed bit widths, RAAD achieves higher accuracy with fewer parameters.


\textbf{Qualitative Results}
Figure~\ref{fig_vis} presents the qualitative results of RAAD on the MVTec-AD and MVTec LOCO datasets. We have visualized the anomaly maps at different stages. Within each group, from left to right, are the anomaly image, ground-truth, predicted anomaly score from EfficientAD-S~\cite{batzner2024efficientad}, predicted anomaly score from after model quantization (with the quantization bit set to 4-bit to highlight the differences), and the anomaly maps generated by RAAD. It is evident that the anomaly maps after quantization exhibit significant diffusion. The anomaly maps produced by RAAD are more accurate than the baseline, with lower anomaly probabilities in the normal regions.

\section{Conclusion}
Unsupervised IAD methods generally suffer from intrinsic bias in normal samples, which results in misaligned attention.
This bias causes models to focus on variable regions while overlooking potential defects in invariant areas.
In response, to this work, we propose RAAD (Recalibrating Attention of Industrial Anomaly Detection), a comprehensive framework that decomposes and recalibrates attention maps through a two-stage quantization process. 
By employing the Hierarchical Quantization Scoring (HQS) mechanism, RAAD optimally redistributes computational resources to enhance defect sensitivity. 
Qualitative and quantitative experiments show that our method can allocate the model attention properly, breaking the bias of unsupervised IAD, and achieving effective attention redistribution. 

{
    \small
    \bibliographystyle{ieeenat_fullname}
    \bibliography{main}

\begin{thebibliography}{45}
\providecommand{\natexlab}[1]{#1}
\providecommand{\url}[1]{\texttt{#1}}
\expandafter\ifx\csname urlstyle\endcsname\relax
  \providecommand{\doi}[1]{doi: #1}\else
  \providecommand{\doi}{doi: \begingroup \urlstyle{rm}\Url}\fi

\bibitem[Batzner et~al.(2024)Batzner, Heckler, and K{\"o}nig]{batzner2024efficientad}
Kilian Batzner, Lars Heckler, and Rebecca K{\"o}nig.
\newblock Efficientad: Accurate visual anomaly detection at millisecond-level latencies.
\newblock In \emph{Proceedings of the IEEE/CVF Winter Conference on Applications of Computer Vision}, pages 128--138, 2024.

\bibitem[Bergmann et~al.(2019{\natexlab{a}})Bergmann, Fauser, Sattlegger, and Steger]{Bergmann2019MVTecAD}
Paul Bergmann, Michael Fauser, David Sattlegger, and Carsten Steger.
\newblock Mvtec ad — a comprehensive real-world dataset for unsupervised anomaly detection.
\newblock In \emph{2019 IEEE/CVF Conference on Computer Vision and Pattern Recognition (CVPR)}, 2019{\natexlab{a}}.

\bibitem[Bergmann et~al.(2019{\natexlab{b}})Bergmann, Löwe, Fauser, Sattlegger, and Steger]{Bergmann2019Improving}
Paul Bergmann, Sindy Löwe, Michael Fauser, David Sattlegger, and Carsten Steger.
\newblock Improving unsupervised defect segmentation by applying structural similarity to autoencoders.
\newblock In \emph{Proceedings of the 14th International Joint Conference on Computer Vision, Imaging and Computer Graphics Theory and Applications}, 2019{\natexlab{b}}.

\bibitem[Bergmann et~al.(2020)Bergmann, Fauser, Sattlegger, and Steger]{Bergmann2020Uninformed}
Paul Bergmann, Michael Fauser, David Sattlegger, and Carsten Steger.
\newblock Uninformed students: Student-teacher anomaly detection with discriminative latent embeddings.
\newblock In \emph{2020 IEEE/CVF Conference on Computer Vision and Pattern Recognition (CVPR)}, 2020.

\bibitem[Bergmann et~al.(2021)Bergmann, Batzner, Fauser, Sattlegger, and Steger]{Bergmann2021MVTecAD}
Paul Bergmann, Kilian Batzner, Michael Fauser, David Sattlegger, and Carsten Steger.
\newblock The mvtec anomaly detection dataset: A comprehensive real-world dataset for unsupervised anomaly detection.
\newblock \emph{International Journal of Computer Vision}, page 1038–1059, 2021.

\bibitem[Bergmann et~al.(2022)Bergmann, Batzner, Fauser, Sattlegger, and Steger]{bergmann2022beyond}
Paul Bergmann, Kilian Batzner, Michael Fauser, David Sattlegger, and Carsten Steger.
\newblock Beyond dents and scratches: Logical constraints in unsupervised anomaly detection and localization.
\newblock \emph{International Journal of Computer Vision}, 130\penalty0 (4):\penalty0 947--969, 2022.

\bibitem[Cai et~al.(2020)Cai, Yao, Dong, Gholami, Mahoney, and Keutzer]{Cai2020ZeroQ}
Yaohui Cai, Zhewei Yao, Zhen Dong, Amir Gholami, MichaelW. Mahoney, and Kurt Keutzer.
\newblock Zeroq: A novel zero shot quantization framework.
\newblock \emph{Cornell University - arXiv,Cornell University - arXiv}, 2020.

\bibitem[Chen et~al.(2023)Chen, Xie, Liu, Wang, Luo, Wang, and Zheng]{chen2023easynet}
Ruitao Chen, Guoyang Xie, Jiaqi Liu, Jinbao Wang, Ziqi Luo, Jinfan Wang, and Feng Zheng.
\newblock Easynet: An easy network for 3d industrial anomaly detection.
\newblock In \emph{Proceedings of the 31st ACM International Conference on Multimedia}, pages 7038--7046, 2023.

\bibitem[Cho(2024)]{cho2024cnn}
Hyuntae Cho.
\newblock Cnn-based autoencoder and post-training quantization for on-device anomaly detection of cartesian coordinate robots.
\newblock In \emph{2024 IEEE 14th Annual Computing and Communication Workshop and Conference (CCWC)}, pages 0662--0666. IEEE, 2024.

\bibitem[Cohen and Hoshen(2020)]{Cohen2020subimage}
Niv Cohen and Yedid Hoshen.
\newblock Sub-image anomaly detection with deep pyramid correspondences.
\newblock \emph{Cornell University - arXiv,Cornell University - arXiv}, 2020.

\bibitem[Deng and Li(2022)]{Deng2022Anomaly}
Hanqiu Deng and Xingyu Li.
\newblock Anomaly detection via reverse distillation from one-class embedding.
\newblock In \emph{2022 IEEE/CVF Conference on Computer Vision and Pattern Recognition (CVPR)}, 2022.

\bibitem[Dong et~al.(2020)Dong, Yao, Arfeen, Gholami, Mahoney, and Keutzer]{dong2020hawq}
Zhen Dong, Zhewei Yao, Daiyaan Arfeen, Amir Gholami, Michael~W Mahoney, and Kurt Keutzer.
\newblock Hawq-v2: Hessian aware trace-weighted quantization of neural networks.
\newblock \emph{Advances in neural information processing systems}, 33:\penalty0 18518--18529, 2020.

\bibitem[Duan et~al.(2023)Duan, Hong, Niu, and Zhang]{duan2023few}
Yuxuan Duan, Yan Hong, Li Niu, and Liqing Zhang.
\newblock Few-shot defect image generation via defect-aware feature manipulation.
\newblock In \emph{Proceedings of the AAAI Conference on Artificial Intelligence}, pages 571--578, 2023.

\bibitem[Esser et~al.(2019)Esser, McKinstry, Bablani, Appuswamy, and Modha]{esser2019learned}
Steven~K Esser, Jeffrey~L McKinstry, Deepika Bablani, Rathinakumar Appuswamy, and Dharmendra~S Modha.
\newblock Learned step size quantization.
\newblock \emph{arXiv preprint arXiv:1902.08153}, 2019.

\bibitem[Gong et~al.(2019)Gong, Liu, Jiang, Li, Hu, Lin, Yu, and Yan]{gong2019differentiable}
Ruihao Gong, Xianglong Liu, Shenghu Jiang, Tianxiang Li, Peng Hu, Jiazhen Lin, Fengwei Yu, and Junjie Yan.
\newblock Differentiable soft quantization: Bridging full-precision and low-bit neural networks.
\newblock In \emph{Proceedings of the IEEE/CVF international conference on computer vision}, pages 4852--4861, 2019.

\bibitem[Haselmann et~al.(2018)Haselmann, Gruber, and Tabatabai]{Haselmann2018Anomaly}
Matthias Haselmann, Dieter~P. Gruber, and Paul Tabatabai.
\newblock Anomaly detection using deep learning based image completion.
\newblock In \emph{2018 17th IEEE International Conference on Machine Learning and Applications (ICMLA)}, 2018.

\bibitem[Huang et~al.(2017)Huang, Liu, Van Der~Maaten, and Weinberger]{huang2017densely}
Gao Huang, Zhuang Liu, Laurens Van Der~Maaten, and Kilian~Q Weinberger.
\newblock Densely connected convolutional networks.
\newblock In \emph{Proceedings of the IEEE conference on computer vision and pattern recognition}, pages 4700--4708, 2017.

\bibitem[Jena et~al.(2024)Jena, Pulkit, Singh, Banerjee, Joshi, Ganesh, Singh, and Bhavsar]{jena2024unified}
Sushovan Jena, Arya Pulkit, Kajal Singh, Anoushka Banerjee, Sharad Joshi, Ananth Ganesh, Dinesh Singh, and Arnav Bhavsar.
\newblock Unified anomaly detection methods on edge device using knowledge distillation and quantization.
\newblock \emph{arXiv preprint arXiv:2407.02968}, 2024.

\bibitem[Kwon et~al.(2019)Kwon, Kim, Kim, Suh, Kim, and Kim]{kwon2019survey}
Donghwoon Kwon, Hyunjoo Kim, Jinoh Kim, Sang~C Suh, Ikkyun Kim, and Kuinam~J Kim.
\newblock A survey of deep learning-based network anomaly detection.
\newblock \emph{Cluster Computing}, 22:\penalty0 949--961, 2019.

\bibitem[Li et~al.(2023)Li, Hu, Li, Chen, Zheng, and Shen]{li2023target}
Hanxi Li, Jianfei Hu, Bo Li, Hao Chen, Yongbin Zheng, and Chunhua Shen.
\newblock Target before shooting: Accurate anomaly detection and localization under one millisecond via cascade patch retrieval.
\newblock \emph{arXiv preprint arXiv:2308.06748}, 2023.

\bibitem[Li et~al.(2021)Li, Gong, Tan, Yang, Hu, Zhang, Yu, Wang, and Gu]{Li2021BRECQ}
Yuhang Li, Ruihao Gong, Xu Tan, Yang Yang, Peng Hu, Qi Zhang, Fei Yu, Wei Wang, and Shi Gu.
\newblock Brecq: Pushing the limit of post-training quantization by block reconstruction.
\newblock \emph{Cornell University - arXiv,Cornell University - arXiv}, 2021.

\bibitem[Liu et~al.(2020)Liu, Li, Zheng, Karanam, Wu, Bhanu, Radke, and Camps]{Liu2020Towards}
Wenqian Liu, Runze Li, Meng Zheng, Srikrishna Karanam, Ziyan Wu, Bir Bhanu, Richard~J. Radke, and Octavia Camps.
\newblock Towards visually explaining variational autoencoders.
\newblock In \emph{2020 IEEE/CVF Conference on Computer Vision and Pattern Recognition (CVPR)}, 2020.

\bibitem[Liu et~al.(2023)Liu, Zhou, Xu, and Wang]{liu2023simplenet}
Zhikang Liu, Yiming Zhou, Yuansheng Xu, and Zilei Wang.
\newblock Simplenet: A simple network for image anomaly detection and localization.
\newblock In \emph{Proceedings of the IEEE/CVF Conference on Computer Vision and Pattern Recognition}, pages 20402--20411, 2023.

\bibitem[Lu et~al.(2023)Lu, Yao, Fu, and Jia]{lu2023removing}
Fanbin Lu, Xufeng Yao, Chi-Wing Fu, and Jiaya Jia.
\newblock Removing anomalies as noises for industrial defect localization.
\newblock In \emph{Proceedings of the IEEE/CVF International Conference on Computer Vision}, pages 16166--16175, 2023.

\bibitem[Ma et~al.(2023)Ma, Jin, Zheng, Wang, Li, Wu, Jiang, Zhang, and Ji]{ma2023ompq}
Yuexiao Ma, Taisong Jin, Xiawu Zheng, Yan Wang, Huixia Li, Yongjian Wu, Guannan Jiang, Wei Zhang, and Rongrong Ji.
\newblock Ompq: Orthogonal mixed precision quantization.
\newblock In \emph{Proceedings of the AAAI conference on artificial intelligence}, pages 9029--9037, 2023.

\bibitem[Nagel et~al.(2019)Nagel, Baalen, Blankevoort, and Welling]{Nagel2019Data}
Markus Nagel, Mart~Van Baalen, Tijmen Blankevoort, and Max Welling.
\newblock Data-free quantization through weight equalization and bias correction.
\newblock In \emph{2019 IEEE/CVF International Conference on Computer Vision (ICCV)}, 2019.

\bibitem[Rezende and Mohamed(2015)]{Rezende2015Variational}
DaniloJimenez Rezende and Shakir Mohamed.
\newblock Variational inference with normalizing flows.
\newblock \emph{International Conference on Machine Learning,International Conference on Machine Learning}, 2015.

\bibitem[Ristea et~al.(2022)Ristea, Madan, Ionescu, Nasrollahi, Khan, Moeslund, and Shah]{Ristea2022Self}
Nicolae-Catalin Ristea, Neelu Madan, Radu~Tudor Ionescu, Kamal Nasrollahi, Fahad~Shahbaz Khan, Thomas~B. Moeslund, and Mubarak Shah.
\newblock Self-supervised predictive convolutional attentive block for anomaly detection.
\newblock In \emph{2022 IEEE/CVF Conference on Computer Vision and Pattern Recognition (CVPR)}, 2022.

\bibitem[Roth et~al.(2022)Roth, Pemula, Zepeda, Sch{\"o}lkopf, Brox, and Gehler]{roth2022towards}
Karsten Roth, Latha Pemula, Joaquin Zepeda, Bernhard Sch{\"o}lkopf, Thomas Brox, and Peter Gehler.
\newblock Towards total recall in industrial anomaly detection.
\newblock In \emph{Proceedings of the IEEE/CVF conference on computer vision and pattern recognition}, pages 14318--14328, 2022.

\bibitem[Rudolph et~al.(2021)Rudolph, Wandt, and Rosenhahn]{Rudolph2021same}
Marco Rudolph, Bastian Wandt, and Bodo Rosenhahn.
\newblock Same same but differnet: Semi-supervised defect detection with normalizing flows.
\newblock In \emph{2021 IEEE Winter Conference on Applications of Computer Vision (WACV)}, 2021.

\bibitem[Ruff et~al.(2021)Ruff, Kauffmann, Vandermeulen, Montavon, Samek, Kloft, Dietterich, and Muller]{Ruff2021a}
Lukas Ruff, Jacob~R. Kauffmann, Robert~A. Vandermeulen, Gregoire Montavon, Wojciech Samek, Marius Kloft, Thomas~G. Dietterich, and Klaus-Robert Muller.
\newblock A unifying review of deep and shallow anomaly detection.
\newblock \emph{Proceedings of the IEEE}, page 756–795, 2021.

\bibitem[Sharmila and Nagapadma(2023)]{sharmila2023quantized}
BS Sharmila and Rohini Nagapadma.
\newblock Quantized autoencoder (qae) intrusion detection system for anomaly detection in resource-constrained iot devices using rt-iot2022 dataset.
\newblock \emph{Cybersecurity}, 6\penalty0 (1):\penalty0 41, 2023.

\bibitem[Sohn et~al.(2020)Sohn, Li, Yoon, Jin, and Pfister]{Sohn2020learning}
Kihyuk Sohn, Chun-Liang Li, Jinsung Yoon, Minho Jin, and Tomas Pfister.
\newblock Learning and evaluating representations for deep one-class classification.
\newblock \emph{Learning,Learning}, 2020.

\bibitem[Sydney et~al.(2019)Sydney, (CMCRC)), (QCRI), and HBKU)]{Sydney2019deep}
RaghavendraChalapathy(Universityof Sydney, CapitalMarketsCooperativeResearchCentre (CMCRC)), SanjayChawla(QatarComputingResearchInstitute (QCRI), and HBKU) HBKU).
\newblock Deep learning for anomaly detection: A survey.
\newblock \emph{arXiv preprint arXiv:2308.06748}, 2019.

\bibitem[Tao et~al.(2022)Tao, Gong, Zhang, Yan, and Adak]{tao2022deep}
Xian Tao, Xinyi Gong, Xin Zhang, Shaohua Yan, and Chandranath Adak.
\newblock Deep learning for unsupervised anomaly localization in industrial images: A survey.
\newblock \emph{IEEE Transactions on Instrumentation and Measurement}, 71:\penalty0 1--21, 2022.

\bibitem[Xie et~al.(2024)Xie, Wang, Liu, Lyu, Liu, Wang, Zheng, and Jin]{xie2024imiad}
Guoyang Xie, Jinbao Wang, Jiaqi Liu, Jiayi Lyu, Yong Liu, Chengjie Wang, Feng Zheng, and Yaochu Jin.
\newblock Im-iad: Industrial image anomaly detection benchmark in manufacturing.
\newblock \emph{IEEE Transactions on Cybernetics}, 54\penalty0 (5):\penalty0 2720--2733, 2024.

\bibitem[Yan et~al.(2022)Yan, Zhang, Xu, Hu, and Heng]{Yan2022Learning}
Xudong Yan, Huaidong Zhang, Xuemiao Xu, Xiaowei Hu, and Pheng-Ann Heng.
\newblock Learning semantic context from normal samples for unsupervised anomaly detection.
\newblock \emph{Proceedings of the AAAI Conference on Artificial Intelligence}, page 3110–3118, 2022.

\bibitem[Yao et~al.(2023)Yao, Li, Qian, Luo, and Zhang]{yao2023focus}
Xincheng Yao, Ruoqi Li, Zefeng Qian, Yan Luo, and Chongyang Zhang.
\newblock Focus the discrepancy: Intra-and inter-correlation learning for image anomaly detection.
\newblock In \emph{Proceedings of the IEEE/CVF International Conference on Computer Vision}, pages 6803--6813, 2023.

\bibitem[You et~al.(2022)You, Cui, Shen, Yang, Lu, Zheng, and Le]{you2022unified}
Zhiyuan You, Lei Cui, Yujun Shen, Kai Yang, Xin Lu, Yu Zheng, and Xinyi Le.
\newblock A unified model for multi-class anomaly detection.
\newblock \emph{Advances in Neural Information Processing Systems}, 35:\penalty0 4571--4584, 2022.

\bibitem[Zagoruyko and Komodakis(2016)]{Zagoruyko2016Wide}
Sergey Zagoruyko and Nikos Komodakis.
\newblock Wide residual networks.
\newblock In \emph{Procedings of the British Machine Vision Conference 2016}, 2016.

\bibitem[Zavrtanik et~al.(2021{\natexlab{a}})Zavrtanik, Kristan, and Skočaj]{Zavrtanik2021DR}
Vitjan Zavrtanik, Matej Kristan, and Danijel Skočaj.
\newblock Dr{-AE}m -- a discriminatively trained reconstruction embedding for surface anomaly detection.
\newblock \emph{arXiv: Computer Vision and Pattern Recognition,arXiv: Computer Vision and Pattern Recognition}, 2021{\natexlab{a}}.

\bibitem[Zavrtanik et~al.(2021{\natexlab{b}})Zavrtanik, Kristan, and Skočaj]{Zavrtanik2021Reconstruction}
Vitjan Zavrtanik, Matej Kristan, and Danijel Skočaj.
\newblock Reconstruction by inpainting for visual anomaly detection.
\newblock \emph{Pattern Recognition}, page 107706, 2021{\natexlab{b}}.

\bibitem[Zhang et~al.(2024)Zhang, Suganuma, and Okatani]{zhang2024contextual}
Jie Zhang, Masanori Suganuma, and Takayuki Okatani.
\newblock Contextual affinity distillation for image anomaly detection.
\newblock In \emph{Proceedings of the IEEE/CVF Winter Conference on Applications of Computer Vision}, pages 149--158, 2024.

\bibitem[Zhang et~al.(2023)Zhang, Li, Li, Dai, Jiang, and Xia]{zhang2023unsupervised}
Xinyi Zhang, Naiqi Li, Jiawei Li, Tao Dai, Yong Jiang, and Shu-Tao Xia.
\newblock Unsupervised surface anomaly detection with diffusion probabilistic model.
\newblock In \emph{Proceedings of the IEEE/CVF International Conference on Computer Vision}, pages 6782--6791, 2023.

\bibitem[Zou et~al.(2022)Zou, Jeong, Pemula, Zhang, and Dabeer]{zou2022spot}
Yang Zou, Jongheon Jeong, Latha Pemula, Dongqing Zhang, and Onkar Dabeer.
\newblock Spot-the-difference self-supervised pre-training for anomaly detection and segmentation.
\newblock In \emph{European Conference on Computer Vision}, pages 392--408. Springer, 2022.

\end{thebibliography}
}


\end{document}